%% file: root.tex
\documentclass[letterpaper, 10pt, conference]{ieeeconf}  

\IEEEoverridecommandlockouts    
\pdfoutput=1

\usepackage{graphics}           
\usepackage{epsfig}             
\usepackage{times}              
\usepackage{amsmath}            
\usepackage{amssymb}            
\usepackage{balance}            
\usepackage{cite}               
\usepackage{graphicx}           
\usepackage{physics}            
\usepackage{xcolor}             
\usepackage{tikz}               

\usepackage{xcolor}
\definecolor{linkblue}{RGB}{0,76,153}

\usepackage{hyperref}           
\hypersetup{
    colorlinks=true,
    linkcolor=linkblue,
    citecolor=black,
    urlcolor=linkblue
}
\usepackage[compatibility=false]{caption}
\usepackage{subcaption}         
\usepackage{lipsum}             
\usepackage{booktabs}           
\usepackage{tabularx}           
\usepackage{wrapfig}            
\usepackage{multirow}           
\usepackage[section]{placeins}  
\usepackage{siunitx}            

\title{\LARGE \bf
Tac4Loco: Learning Spatiotemporal Plantar Pressure Representations for Humanoid Locomotion
}

\author{
Ziyun Liu$^{1}$, Sikai Guo$^{1}$, Zheng Li$^{1}$, Jiahang Cao$^{2}$, Haichao Liu$^{3}$, Pei Qu$^{1}$, Yinghong Zhang$^{1}$, \\
Jinni Zhou$^{1}$, Jun Ma$^{1\dag}$%
\thanks{$^{1}$The Hong Kong University of Science and Technology (Guangzhou).
        {\tt\small \{zliu176, sguo837, zli514, pqu458, yzhang780\}@connect.hkust-gz.edu.cn; eejinni@hkust-gz.edu.cn; jun.ma@ust.hk}}%
\thanks{$^{2}$The University of Hong Kong.
        {\tt\small jiahang@connect.hku.hk}}
\thanks{$^{3}$Nanyang Technological University.
        {\tt\small haichao.liu@ntu.edu.sg}}
        \thanks{\dag Corresponding Author.}
}

\begin{document}
\maketitle

\input{Sections/0_Abstract}

\input{Sections/1_Introduction}

\input{Sections/2_RelatedWorks}

\input{Sections/4_Methodology}
\input{Sections/5_Experiments}
\input{Sections/6_Conclusion}


\input{Sections/9_Bibliography}

\end{document}

%% file: Sections/0_Abstract.tex
\begin{abstract}
    Humanoid robots are expected to traverse complex terrains, where the plantar support may vary dramatically due to foot placement errors, ground properties, and transient dynamics. 
    To achieve robust locomotion, the robots are required to adapt to uneven terrain and uncertain foot--ground interactions. 
    Existing locomotion policies rely primarily on proprioception or exteroceptive terrain perception, where the former provides only indirect evidence of plantar support, while the latter predicts contact conditions before touchdown but cannot observe the actual support in real-time. 
    Although some studies incorporate plantar contacts as an auxiliary perception, they rely mainly on summary statistics, overlooking the spatial topology of plantar pressure, which provides a more direct characterization of the realized contact state. 
    To bridge this gap, we present Tac4Loco, a tactile-perceptive framework that incorporates multi-array plantar pressure as direct feedback for humanoid locomotion. 
    We formulate a topology-preserving ordinal representation to map simulated and physical sensor signals into a shared observation space,
    with a dual-branch encoder for extracting their spatial and temporal representations. Subsequently, the learned spatiotemporal features are integrated with augmented proprioception including terrain estimation cues, and provided to an asymmetric actor-critic architecture for policy learning.
    Extensive simulation and real-world experiments demonstrate improved  tracking performance and support adaptation on terrains with inclined, partial, asymmetric, and changing support. We further demonstrate its zero-shot deployment on unseen compliant and unstructured terrains, including a foam platform and a gravel road. All code and experimental configurations will be released as open-source to facilitate reproducibility.

\end{abstract}

%% file: Sections/1_Introduction.tex
\section{Introduction} \label{sec 1}

    Humanoid robots have the potential to traverse complex, human-centered environments containing uneven ground, terrain boundaries, slopes, and other irregular surfaces~\cite{2026RoboGauge,2026planc,2026nowYouSeeThat,2026fastStair,2025hugWBC,2024real-world,2025vb-com,2026rpl,2025pim}. 
    Although terrain geometry constrains possible footholds, the support realized beneath each foot remains uncertain. For instance, a foot may land on an inclined surface, receive only partial support near an edge, or experience asymmetric medial--lateral loading because of placement errors, surface compliance, friction, or transient dynamics. These contact conditions further evolve with gait phase, foot placement, and body motion, directly affecting locomotion stability.
    
    Existing locomotion policies primarily rely on proprioceptive adaptation or exteroceptive terrain perception. 
    Blind methods improve robustness through domain randomization~\cite{2024real-world}, curriculum learning~\cite{2025hugWBC,2026RoboGauge,2024toward}, and temporal policy architectures~\cite{2026hord,2024toward,2024real-world}, but plantar support remains only implicitly inferred from proprioception, which is insufficient and indirect. 
    Perceptive methods leverage depth images or elevation maps for terrain preview, foothold selection, and anticipatory body adaptation, but such geometric priors only estimate the contact conditions before touchdown, without direct feedback on whether the feet are supported by the ground.
    
    Foot-mounted sensors can measure these contact outcomes directly, but their signals are often reduced to binary contact states, total ground reaction forces, centers of pressure, or low-dimensional tactile features~\cite{2025lipschitz,2022robust-contact}. 
    Such summaries discard the spatial topology of plantar support, including contact shape, local pressure peaks, edge loading, and fore--aft or medial--lateral load redistribution. 
    For example, full-sole contact and two localized contact patches can produce comparable total forces and a similar center of pressure, despite having different support patterns. 
    Distinguishing such contact configurations therefore requires a multi-array, topology-preserving plantar pressure representation that captures both the spatial load distribution and its temporal evolution.
    \begin{figure}
        \centering
        \vspace{6pt}
        \includegraphics[width=\linewidth]{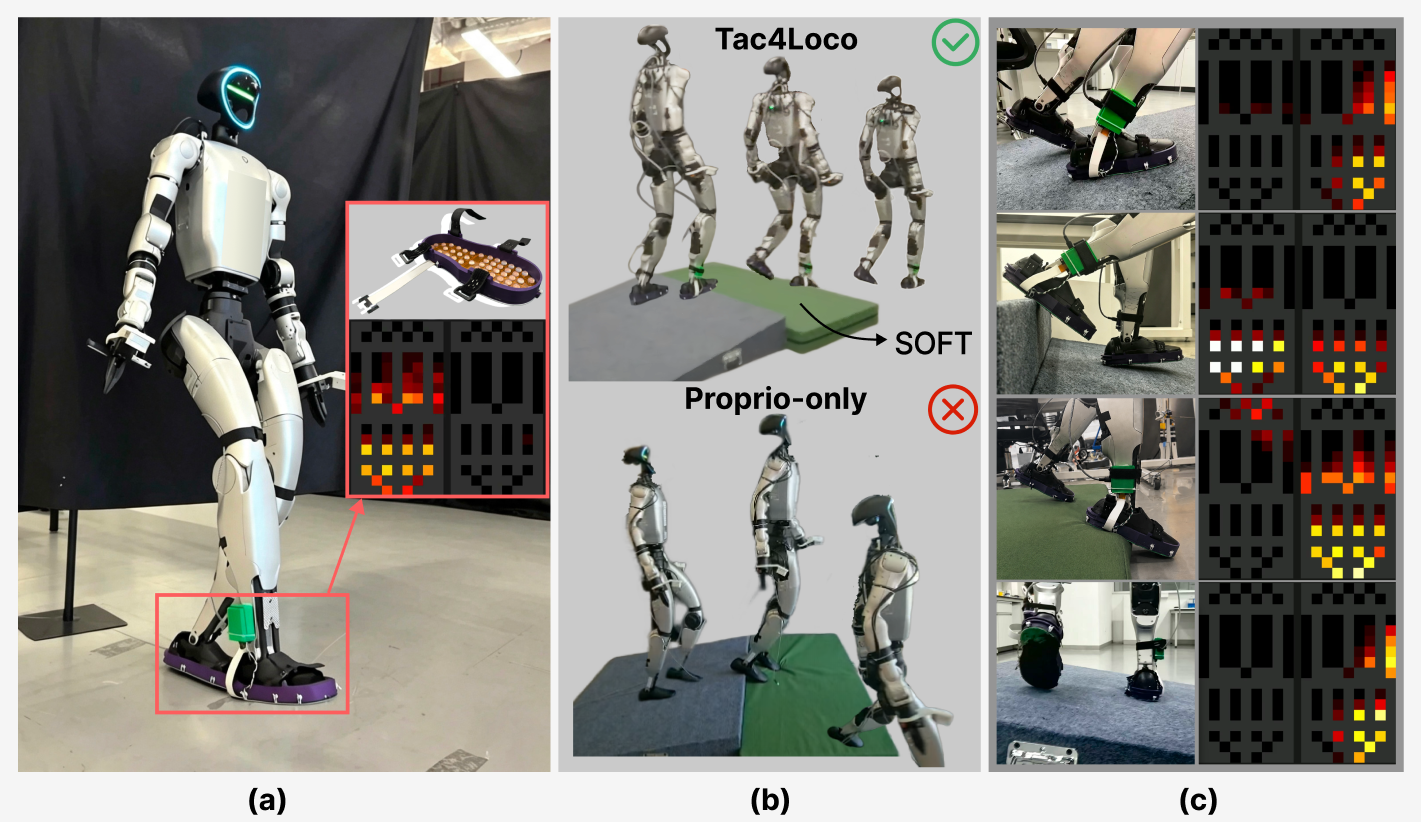}
        \caption{Real-world deployment of Tac4Loco with plantar pressure feedback.
        (a) Unitree G1 equipped with bilateral plantar pressure insoles.
        (b) Representative comparison on a compliant terrain transition, where Tac4Loco succeeds while the proprioception-only baseline fails.
        (c) Selected support states and their corresponding plantar pressure maps, highlighting distinct pressure topologies under different contact conditions.}
        \label{fig:teaser}
    \end{figure}
    
    To address this gap, we propose Tac4Loco, a tactile-perceptive humanoid locomotion framework that augments proprioception with multi-array plantar pressure as direct post-contact feedback of foot--ground interaction.
    Tac4Loco constructs a topology-preserving pressure matrix for each foot by mapping simulated contact forces and physical sensor measurements into shared ordinal load bins, thereby alleviating the sim-to-real gap in pressure sensing.
    To provide a compact physical cue about the local support, it derives contact-conditioned terrain-orientation features from proprioceptive kinematics and pressure-based contact confidence, forming an augmented proprioception.
    To capture both instantaneous support topology and temporal load transfer, a dual-branch encoder combines a state-conditioned spatial encoder with a temporal pressure--proprioceptive encoder.
    The resulting tactile representations and augmented proprioception are fused within an asymmetric actor--critic framework for end-to-end reinforcement learning (RL).
    In simulation, Tac4Loco improves locomotion performance and support adaptation over a proprioception-only baseline across several challenging terrains. As shown in Fig.~\ref{fig:teaser}, real-world experiments further evaluate its deployability under inclined, partial, asymmetric, and changing support, including compliant conditions absent from training.
    
    Our contributions are as follows:

    \begin{itemize}
        \item We propose Tac4Loco, a deployable learned humanoid locomotion policy utilizing spatiotemporal plantar pressure as direct post-contact feedback to complement proprioception under uncertain terrain conditions.
        
        \item We design a topology-preserving ordinal quantization scheme to bridge the sim-to-real gap for tactile sensing, alongside a novel dual-branch architecture that explicitly decouples spatiotemporal representation.
    
        \item Through extensive simulation and physical hardware experiments on a Unitree G1 humanoid robot, we demonstrate enhanced tracking precision, heading regulation, and zero-shot transfer to unseen compliant and granular terrains.
    
    \end{itemize}

%% file: Sections/2_RelatedWorks.tex
\section{Related Work}\label{sec 2}
    \subsection{Proprioceptive Humanoid Locomotion}        
        Proprioceptive locomotion avoids the latency and visibility limitations of exteroceptive sensing and enables robust real-world humanoid control. Representative studies use causal history modeling for zero-shot sim-to-real transfer~\cite{2024real-world} and validate proprioception-only policies~\cite{2025vb-com}. 
        Recent studies introduce more complementary strategies for improving the robustness. Wang et al.~\cite{2024toward} identify critical estimated states, and Gu et al.~\cite{2024mastering} learn compact latent representations via world models. HoRD~\cite{2026hord} improves adaptation with history-conditioned reinforcement learning and distillation, Chen et al.~\cite{2025lipschitz} enhance smoothness and transferability via Lipschitz regularization, while HugWBC~\cite{2025hugWBC} enables unified whole-body tracking. These methods show that temporal modeling and policy regularization improve robustness under uncertainty.
        However, proprioceptive methods remain inherently reactive, inferring terrain and contact conditions only after they are reflected in the robot, which decreases the stability during humanoid locomotion.

    \subsection{Perceptive Humanoid Locomotion}
        Compared with proprioceptive methods, perceptive methods incorporate additional sensing modalities to alleviate partial observability issues. Vision-based methods obtain terrain preview from depth images or elevation maps. Depth-image-driven approaches learn terrain representations or distill policies from noisy depth observations~\cite{2026dpl,2026rpl,2026nowYouSeeThat}, whereas map-based methods construct and encode local elevation representations~\cite{2025pim,2026gaitAdaptive,2026AME2}. Other methods combine perception with planning for foothold selection~\cite{2026planc,2026fastStair}. Nevertheless, vision is vulnerable to occlusion, latency, calibration errors, and depth noise, and cannot verify foot loading after touchdown; locomotion must remain robust to unreliable visual perception~\cite{2025vb-com}. 
        
        Therefore, recent works employ foot-mounted tactile information and pressure sensors to directly measure contact timing, force, and pressure distribution. Guadarrama-Olvera et al.~\cite{2024partialFootholdZMP} reconstruct the support polygon from distributed plantar tactile feedback for model-based stabilization over partial footholds. PressMimic~\cite{lu2026pressmimic} incorporates pressure into motion capture and uses pressure-derived supervision for contact-consistent humanoid imitation, while force-aware objectives regulate impact and ground reaction forces~\cite{2026quietWalk}. However, using multi-array plantar pressure as a deployable observation for learned humanoid locomotion remains underexplored.

%% file: Sections/4_Methodology.tex
\section{Method}\label{sec:method}

\begin{figure*}[t]
        \centering
        \vspace{6pt}
        \includegraphics[width=1\linewidth]{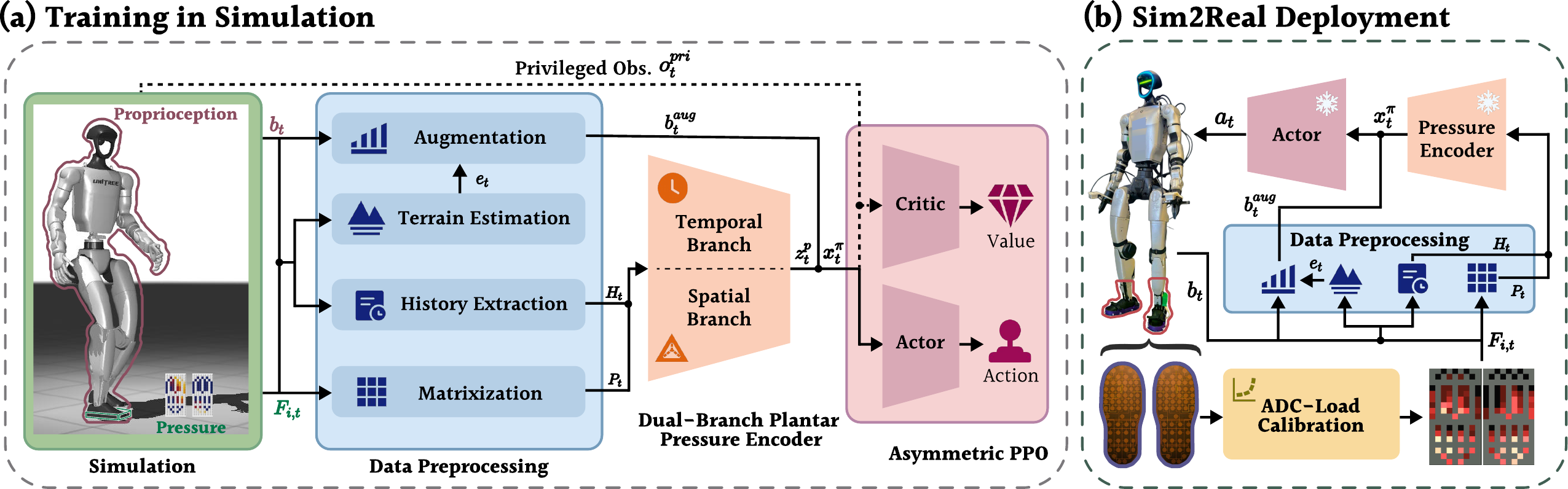}
        \caption{Overview of the Tac4Loco pipeline. During training, proprioception and raw  plantar pressure measurements are collected in simulation. The pressure signals are preprocessed through matrixization, history extraction, terrain-orientation estimation, and observation augmentation, and then encoded by a dual-branch plantar pressure encoder consisting of a spatial branch and a temporal branch. The resulting tactile representation is fused with proprioception in an asymmetric actor--critic framework for policy learning. After training, the learned actor is transferred to the real robot and driven by physical plantar pressure measurements processed under the same representation.}
        \label{fig:pipeline}
    \end{figure*}

    Tac4Loco is an RL-based end-to-end tactile-perceptive humanoid locomotion controller. 
    The key insight of our framework is to treat plantar pressure as posterior information about the realized foot--ground interaction, complementing proprioception with direct feedback of how each foot is actually supported during motion. 
    As shown in Fig.~\ref{fig:pipeline}, Tac4Loco takes proprioception and plantar pressure as inputs, preprocesses the pressure observations, and encodes them into latent representations with a dual-branch encoder, which is subsequently integrated with the augmented proprioception as the policy input. Finally, Tac4Loco learns a pressure-aware locomotion policy through an asymmetric actor--critic framework, and outputs the whole-body residual joint positions. The trained actor is directly deployed on the real robot using physical plantar pressure measurements.

    In this section, we first establish the problem
    definition, providing a rigorous definition of the state and action spaces. Next, we detail the core components of Tac4Loco, highlighting the key contributions of each modules.

    \subsection{Problem Formulation}
    \label{subsec:formulation}
    
    We model pressure-aware humanoid locomotion as a discrete-time Markov Decision Process. At each control step $t$, the actor receives a deployable observation $o_t$ and samples an action $a_t\sim\pi_\theta(\cdot|o_t)$. 

    The deployable observation $o_t=[b_t, P_t, H_t, e_t]$ contains four structured components.
    $b_t$ denotes the deployable proprioceptive observation, including IMU measurements, commanded velocity, joint states, the previous action, and gait phase. $P_t$ denotes the current ordinal plantar pressure observation, while $H_t$ summarizes recent pressure evolution together with compact motion context. The contact-conditioned terrain-orientation feature $e_t$ provides an auxiliary physical cue about the realized local support. The output action $a_t \in \mathbb{R}^{29}$ represents the residual joint-position targets relative to a nominal standing pose.
    
            
    
\subsection{Plantar Pressure Sensing and Quantization}
\label{subsec:pressure_sensing}


In our hardware design, each foot has a 60-element FSR insole aligned one-to-one with the simulated taxels, digitized by a 12-bit ADC at 50 Hz. Because the FSR force–resistance response is nonlinear and element-dependent, identical loads yield systematically different ADC codes. This heterogeneity is intrinsic to FSR arrays rather than temporal measurement noise, so force-domain randomization alone cannot remove it. Absolute force is thus not a stable deployment interface, and per-taxel calibration is impractical. 


To address this, we map both simulated contact forces and calibrated hardware readings into a common force domain and quantize them with a single operator \(Q(\cdot)\). Since locomotion depends mainly on spatial load patterns and their temporal evolution rather than absolute magnitudes, the policy can instead consume ordinal load levels, whose relative structure remains stable under element-wise curve offsets.
Concretely, we measure ADC codes under known calibration loads spanning the pad full scale of \(2000\,\mathrm{g}\) and fit a quadratic ADC-to-load map  \(\hat{m}(a)\):
\begin{equation}
  \hat{m}(a)=c_2 a^2+c_1 a+c_0,
  \qquad
  F_{\mathrm{real}}(a)=\hat{m}(a)\,g_0,
\end{equation}
where \(g_0=9.81\times10^{-3}\,\mathrm{N/g}\), clamped at $\hat{m}(a)=0$ for $a<50$ and $\hat{m}=2000\,\mathrm{g}$ for $a>4000$.
In simulation, \(F_{\mathrm{sim}}\) is the compressive sole-normal component of each taxel contact force, expressed in the same units. 
To account for taxel response variance at higher loads, the quantization operator $Q(F)$ employs non-uniform bin steps: $50\,\mathrm{g}$ resolution within $[0, 1000]\,\mathrm{g}$ and $100\,\mathrm{g}$ resolution within $(1000, 2000]\,\mathrm{g}$, yielding discrete ordinal levels $q_{i,t} = Q(F) \in \{0, \dots, 30\}$. This shared discrete alphabet absorbs inter-element response discrepancies while retaining spatial load gradients.

Applying this quantizer to both domains, $q_{\mathrm{sim}}=Q(F_{\mathrm{sim}})$ and $q_{\mathrm{real}}=Q\!\big(F_{\mathrm{real}}(a)\big)$, yields the ordinal pressure \(q_{i,t}\), a shared observation alphabet for simulation and deployment that absorbs inter-taxel characteristic mismatch.  Training then randomizes the residual dynamic artifacts: force-domain gain and asymmetric hysteresis, zero-order hold, and level-domain perturbations with structured dropouts, all applied before the pressure observations in Sec.~\ref{subsec:pressure_observation} are formed.
    
    \subsection {Plantar Pressure Observation Construction}
    \label{subsec:pressure_observation}
    
    Starting from the ordinal taxel pressures $q_{i,t}$ obtained in
Sec.~\ref{subsec:pressure_sensing}, we construct the spatial map $P_t$ and
the temporal history $H_t$.
    \subsubsection{Spatial Plantar Pressure Observation}
    The quantized values are then arranged according to their physical sensor locations on a topology-preserving plantar grid, preserving the spatial layout of foot--ground loading. The resulting bilateral spatial observation is $P_t=[P_t^L,P_t^R]$, where $P_t^L$ and $P_t^R$ denote the left- and right-foot ordinal pressure maps respectively.
    
    \subsubsection{Temporal Pressure--Proprioceptive History}
    To construct the temporal observation efficiently, each spatial observation $P_t$ is compressed into a 21-dimensional spatial descriptor $\phi(P_t)$. The descriptor contains regional pressure proportions from a $2\times4$ partition of each foot, bilateral contact indicators, the load-balance ratio, and fore--aft and medial--lateral differences between the pressure-derived centers of pressure.
    Subsequently, we fuse $\phi(P_t)$ with a compact proprioceptive context $C_t$ to form the single-frame packet $s_t=[\phi(P_t),C_t]\in\mathbb{R}^{37}$. Here, $C_t\in\mathbb{R}^{16}$ comprises base angular velocity, projected gravity, and bilateral leg joint-state statistics, providing motion context for interpreting pressure patterns across postures and gait phases.

    Finally, we construct the temporal pressure--proprioceptive observation $H_t$ as an eight-packet sequence:
    \begin{equation}
        H_t=[s_{t-\kappa_5},\ldots,s_{t-\kappa_1},s_{t-2},s_{t-1},s_t].
    \end{equation}
    Here, $\kappa_1,\ldots,\kappa_5$ denote five progressively earlier sampling offsets. We adopt a near-dense and far-sparse scheme: the three contiguous recent packets capture rapid touchdown and load-transfer transients, while the five sparsely sampled earlier packets extend the temporal coverage without increasing the sequence length. The resulting history $H_t$ is then provided to the temporal encoder branch.
    
    \subsection{Contact-Conditioned Terrain-Orientation Estimation}
    \label{subsec:support_orientation}
    
    To provide the policy with a compact physical cue about the realized support under each foot, Tac4Loco derives the bilateral terrain-orientation estimates $\theta_t^L,\theta_t^R\in\mathbb{R}^{2}$ from the robot base orientation and leg forward kinematics, representing the fore--aft and medial--lateral sole inclinations in the robot-heading coordinate frame. 
    However, a kinematic sole orientation reflects the actual local support only when the foot is sufficiently planted. Therefore, we introduce the corresponding support confidences $c_t^L$ and $c_t^R$ from the number and spatial spread of active pressure locations, where sparse, localized, or absent contact indicates lower support reliability.
    Based on these confidences, each orientation estimate is updated only under reliable plantar contact; otherwise, the previous estimate is retained while its confidence decays. We further derive an overall caution cue $\eta_t$ from the bilateral confidences to summarize the reliability of the current support information.

    Finally, the orientation estimates and their reliability information form the contact-conditioned terrain-orientation feature:
    \begin{equation}
        e_t =
        [\theta_t^L,\theta_t^R,c_t^L,c_t^R,\eta_t].
    \end{equation}
    To expose this contact-conditioned cue to the downstream encoder and policy, we concatenate $e_t$ with the base proprioceptive observation $b_t$ to form the augmented observation $b_t^{\mathrm{aug}}=[b_t,e_t]$.
    
    \subsection{Dual-Branch Plantar Pressure Encoder}
    \label{subsec:pressureEncoder}

    Tac4Loco jointly encodes the spatial observation $P_t$ and temporal history $H_t$ to obtain complementary tactile features for policy learning. As shown in Fig.~\ref{fig:dual_branch_encoder}, the state-conditioned spatial branch extracts context-relevant instantaneous support topology from $P_t$, while the temporal pressure--proprioceptive branch captures recent support evolution from $H_t$.
    
    \subsubsection{State-conditioned Spatial Branch}
    
    \begin{figure}[t]
        \centering
        \vspace{6pt}
        \includegraphics[width=\linewidth]
        {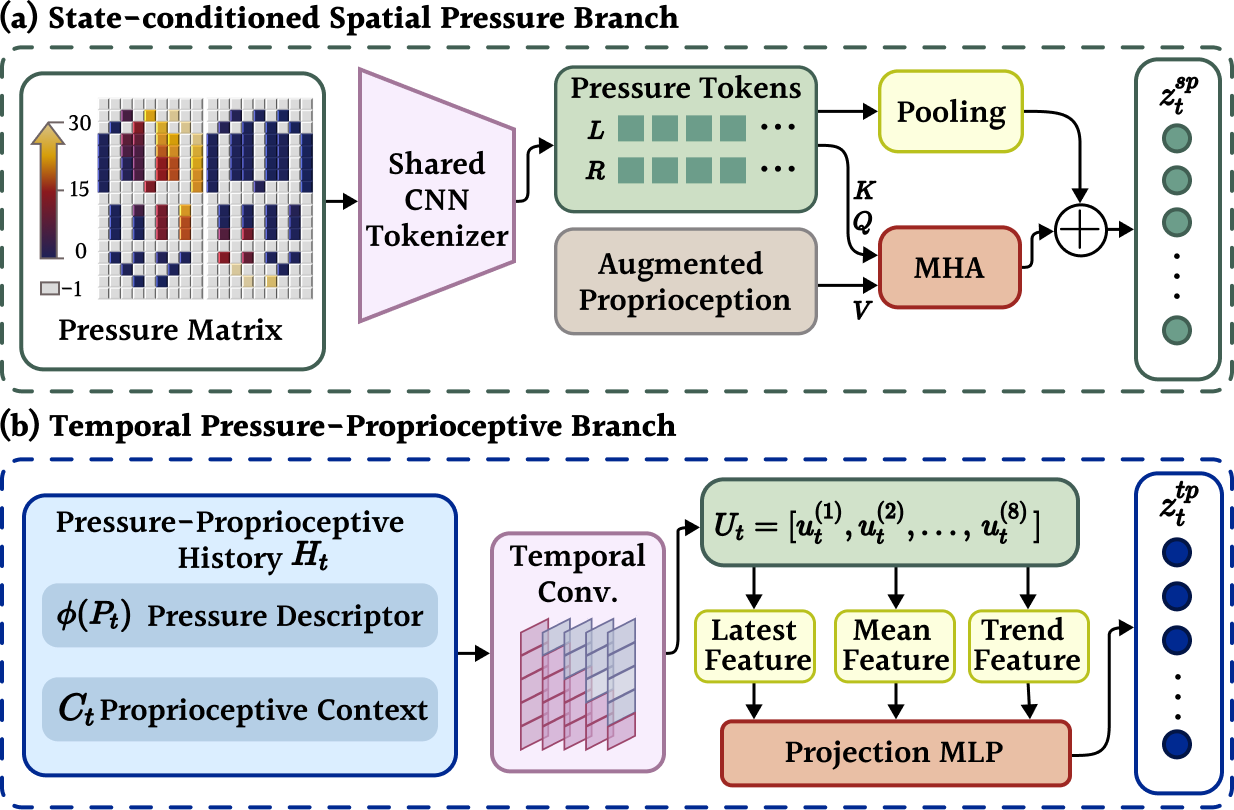}
        \caption{Architecture of the dual-branch plantar pressure encoder, comprising the state-conditioned spatial branch (a) and temporal pressure--proprioceptive branch (b).}
        \label{fig:dual_branch_encoder}
    \end{figure}

    As shown in Fig.~\ref{fig:dual_branch_encoder}(a), first, the ordinal pressure maps $P_t$ are normalized per foot and converted by a shared CNN tokenizer into left- and right-foot token sequences. Each token is then augmented with spatial-position and foot-identity embeddings, and the resulting sequences are combined into the bilateral pressure-token set $T_t^p$ for subsequent attention.

    These tokens encode the instantaneous loading topology, but pressure alone does not determine how that topology should be interpreted under the current motion. For example, the same plantar pattern may indicate expected load transfer in one gait phase, but unstable support in another. Therefore, the augmented observation $b_t^{\mathrm{aug}}$ is projected into a query token to condition the pressure representation on the current velocity command, joint states, gait phase, and support reliability, while $T_t^p$ serves as the spatial keys and values in multi-head attention (MHA)~\cite{vaswani2017attention}:
    \begin{equation}
        z_t^{\mathrm{att}}
        =
        \operatorname{MHA}
        \bigl(
        \phi_q(b_t^{\mathrm{aug}}),
        \phi_k(T_t^p),
        \phi_v(T_t^p)
        \bigr),
    \end{equation}
    where $\phi_q$, $\phi_k$, and $\phi_v$ are learnable projections. This cross-attention allows the current locomotion state to retrieve the most relevant plantar regions.
    
    Finally, $z_t^{\mathrm{att}}$ is fused with mean- and max-pooled token features and projected into the spatial latent $z_t^{sp}$. Attention adaptively selects state-conditioned local support cues, while pooling preserves global and salient patterns, allowing $z_t^{sp}$ to capture contact configurations such as slopes, partial footholds, edge contacts, and asymmetric loading.
    
    \subsubsection{Temporal Pressure--Proprioceptive Branch}
    While the spatial branch captures instantaneous support topology, it cannot describe how contact evolves over time. Therefore, we introduce the temporal pressure--proprioceptive branch to capture recent support evolution. 
    As shown in Fig.~3(b), a one-dimensional temporal convolutional layer~\cite{bai2018empirical} first maps the eight-packet history $H_t$ to a feature sequence $U_t=[u_t^{(1)},\ldots,u_t^{(8)}]$. We then summarize $U_t$ using its latest feature, temporal mean, and latest-to-earliest difference, which characterize the current support state, persistent context, and recent change trend, respectively. Their fused representation is fed into an MLP to obtain the temporal latent $z_t^{tp}$, providing a compact description of recent pressure and proprioceptive changes.
    
    Finally, the spatial and temporal plantar pressure latents are combined as:
    \begin{equation}
        z_t^p=[z_t^{sp},z_t^{tp}].
    \end{equation}

    \subsection{Policy Learning and Training Details}
    \label{subsec:trainingSetup}
    
    The unified plantar pressure representation $z_t^p$ is fused with
    the augmented base observation to form the actor input:
    \begin{equation}
        x_t^\pi =
        \left[
        b_t^{\mathrm{aug}},
        z_t^p
        \right].
    \end{equation}
    The actor maps $x_t^\pi$ to the action distribution.
    
    Tac4Loco adopts an asymmetric actor--critic structure~\cite{pinto2018asymmetric}: the actor uses only deployable observations, while the critic additionally receives privileged simulator states for value estimation, including base linear velocity, foot--ground contact quantities, and (on rough terrain) a local height map, together with an unperturbed plantar-pressure reading.
    
   The policy is optimized using PPO~\cite{schulman2017proximal} with standard locomotion rewards and a terrain curriculum. To improve deployment robustness, training perturbs the plantar pressure observations with sensor-rate variations, sensitivity shifts, noise, and temporary dropouts.
    
    Rewards are organized into four groups.
    \emph{(i)~Task tracking} encourages matching the commanded base linear and yaw velocities.
    \emph{(ii)~Motion regularization} penalizes excessive torso tilt and angular rates, enforces a mode-dependent reference posture, and discourages abrupt actions, joint-limit violations, and early terminations.
    \emph{(iii)~Gait and contact shaping} promotes periodic footfalls with adequate swing clearance, while penalizing foot slip, hard landings, and residual motion under near-zero commands.
    \emph{(iv)~Plantar support alignment} rewards coincidence between the sole normal and the force-weighted ground-contact normal on supporting feet, encouraging the feet to plant flush with the local terrain.
    The reward composition and coefficients, including the plantar support alignment term, are identical across all policy variants.

%% file: Sections/5_Experiments.tex
\section{Experiments}\label{sec:result}
\subsection{Experimental Setup}
\label{subsection:Experimental Setup}
    \subsubsection{Simulation Setup}
    All simulation experiments are conducted in MJLab, a GPU-parallel MuJoCo-based environment, using a Unitree G1 robot model. The policy outputs joint-position targets at 50\,Hz, with low-level PD motor loops at 200\,Hz. To emulate the real plantar array, each foot is equipped with 60 spherical contact geoms placed at the physical taxel locations; these spheres serve as the sole--ground interface. Contact forces on each sphere are read from the simulator and projected onto the local sole normal, yielding a non-negative compressive force that corresponds to plantar pressure.
    
    \subsubsection{Terrain Configuration}
    The simulation terrains include flat ground, gently undulating terrain with an overall height range of 15\,cm, ascending and descending slopes at $15^\circ$ and $20^\circ$, random support-height terrain with height differences up to 15\,cm, and V-shaped trenches formed by two opposing inclined walls with inclinations randomly selected from $0^\circ$ to $25^\circ$.



    \subsubsection{Policy Variants}
    We evaluate three policies compared against Tac4Loco:
    (i)~\textbf{Proprio-only}: the official open-source Unitree proprioceptive locomotion baseline~\cite{unitree_rl_mjlab}, trained with the same reward configuration but without plantar pressure inputs;
    (ii) \textbf{w/o Pressure Enc.}: removes the spatial and temporal pressure latents while retaining the terrain-orientation feature $e_t$;
    (iii) \textbf{w/o Terrain Est.}: retains the dual-branch pressure encoder but removes $e_t$.
    None of the policies receives visual perception.

    \subsubsection{Evaluation Metrics} 
    
    For each terrain and metric, one scalar is first computed per episode. Episode-level values are then pooled over all commands included for that metric. Continuous metrics are reported as mean $\pm$ standard deviation, while survival is reported as a percentage. (a) Task-level metrics include survival rate and linear- and yaw-velocity tracking errors. Survival rate is the fraction of episodes that reach the time horizon without falling or early termination. The linear-velocity tracking error is computed as the mean Euclidean error between the commanded and actual planar velocities over alive frames, while the yaw-velocity tracking error is the mean absolute error between the commanded and actual yaw velocities. (b) Directional regulation is measured by the drift angle between the commanded direction and the actual displacement direction: \[ D_{\mathrm{angle}} = \left| \operatorname{atan2} \left( d_x e_{c,y} - d_y e_{c,x}, \mathbf{d}^{\top}\hat{\mathbf{e}}_c \right) \right|. \] Here, \[ \mathbf{d} = \mathbf{p}_{xy,t_{\mathrm{end}}} - \mathbf{p}_{xy,0}, \] where $t_{\mathrm{end}}$ denotes the last alive frame, and $\hat{\mathbf{e}}_c$ is the commanded translational direction transformed into the world frame according to the initial heading. Episodes with $\|\mathbf{d}\|_2 < 0.5\,\mathrm{m}$ or zero translational commands are excluded from this metric. The drift angle is reported in radians, with smaller values indicating better directional regulation. (c) Gait behavior is characterized by step length. A touchdown is detected at a contact rising edge that persists for at least three control steps. Step length is the mean horizontal distance between consecutive touchdown locations of the same foot, averaged across the two feet, and is interpreted jointly with survival and velocity tracking.
    
\subsection{Simulation Evaluation on Complex Terrains}

\begin{figure}[t]
\centering
\vspace{6pt}
\includegraphics[width=1\columnwidth]{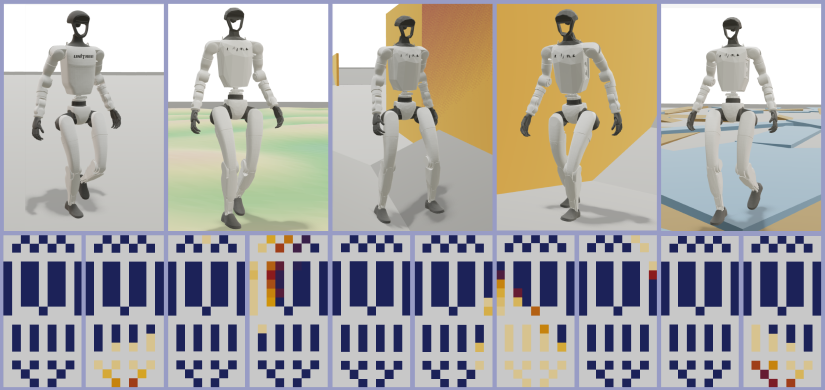}
\caption{Representative simulation terrains and the corresponding bilateral plantar pressure maps of Tac4Loco:
(1)~flat terrain,
(2)~gently undulating terrain,
(3)~V-trench terrain,
(4)~slope terrain,
and
(5)~random support-height terrain.}
\label{fig:sim_terrain}
\end{figure}

As shown in Fig.~\ref{fig:sim_terrain}, representative robot states are paired with their bilateral plantar pressure maps. The examples illustrate full, shifted, localized, and asymmetric support patterns under continuous and discontinuous terrain variations.

\begin{table}[t]
\centering
\caption{Overall simulation performance of Tac4Loco.}
\label{tab:sim_main_results}

\renewcommand{\arraystretch}{1.12}

\resizebox{\columnwidth}{!}{
\begin{tabular}{llccc}
\toprule
Terrain
& Method
& Survival (\%) $\uparrow$
& Lin. Vel. Err. (m/s) $\downarrow$
& Yaw Vel. Err. (rad/s) $\downarrow$ \\
\midrule

FL
& Proprio-only
& 100.0
& 0.337 $\pm$ 0.244
& 0.380 $\pm$ 0.224 \\
& Tac4Loco
& 100.0
& \textbf{0.131 $\pm$ 0.024}
& \textbf{0.329 $\pm$ 0.189} \\

\midrule
GU
& Proprio-only
& 100.0
& 0.381 $\pm$ 0.252
& 0.373 $\pm$ 0.235 \\
& Tac4Loco
& 100.0
& \textbf{0.140 $\pm$ 0.030}
& \textbf{0.339 $\pm$ 0.185} \\

\midrule
RSH
& Proprio-only
& 71.7
& 0.351 $\pm$ 0.248
& 0.416 $\pm$ 0.207 \\
& Tac4Loco
& \textbf{96.5}
& \textbf{0.168 $\pm$ 0.051}
& \textbf{0.378 $\pm$ 0.159} \\

\midrule
SP
& Proprio-only
& 22.0
& 1.018 $\pm$ 0.038
& \textbf{0.238 $\pm$ 0.021} \\
& Tac4Loco
& \textbf{77.9}
& \textbf{0.259 $\pm$ 0.042}
& 0.283 $\pm$ 0.022 \\

\midrule
VT
& Proprio-only
& 100.0
& 0.952 $\pm$ 0.022
& 0.563 $\pm$ 0.006 \\
& Tac4Loco
& 100.0
& \textbf{0.309 $\pm$ 0.239}
& \textbf{0.426 $\pm$ 0.048} \\

\bottomrule
\end{tabular}}

\begin{flushleft}
\footnotesize
\textit{Note.}
FL = flat, GU = gently undulating, RSH = random support-height,
SP = slopes, and VT = V-trench.
Values denote mean $\pm$ standard deviation.
\end{flushleft}

\end{table}

Table~\ref{tab:sim_main_results} reports overall task performance, Table~\ref{tab:drift_angle} evaluates open-terrain directional regulation, and Fig.~\ref{fig:step_length} summarizes step length under challenging support conditions.

\subsubsection{Overall Task Execution}
Tac4Loco maintains 100\% survival on flat, gently undulating, and V-trench terrains, and raises survival from 71.7\% to 96.5\% on random support height and from 22.0\% to 77.9\% on slopes. It also reduces linear-velocity error across all terrain groups, while yaw
tracking is generally improved or comparable except on slopes.

\subsubsection{Directional Regulation}
As shown in Table~\ref{tab:drift_angle}, Tac4Loco consistently reduces drift angle on flat, gently undulating, and random support-height terrains, indicating better preservation of the commanded translational direction over each episode.

\subsubsection{Support Metrics}
\begin{figure}
    \centering
    \includegraphics[width=0.7\linewidth]{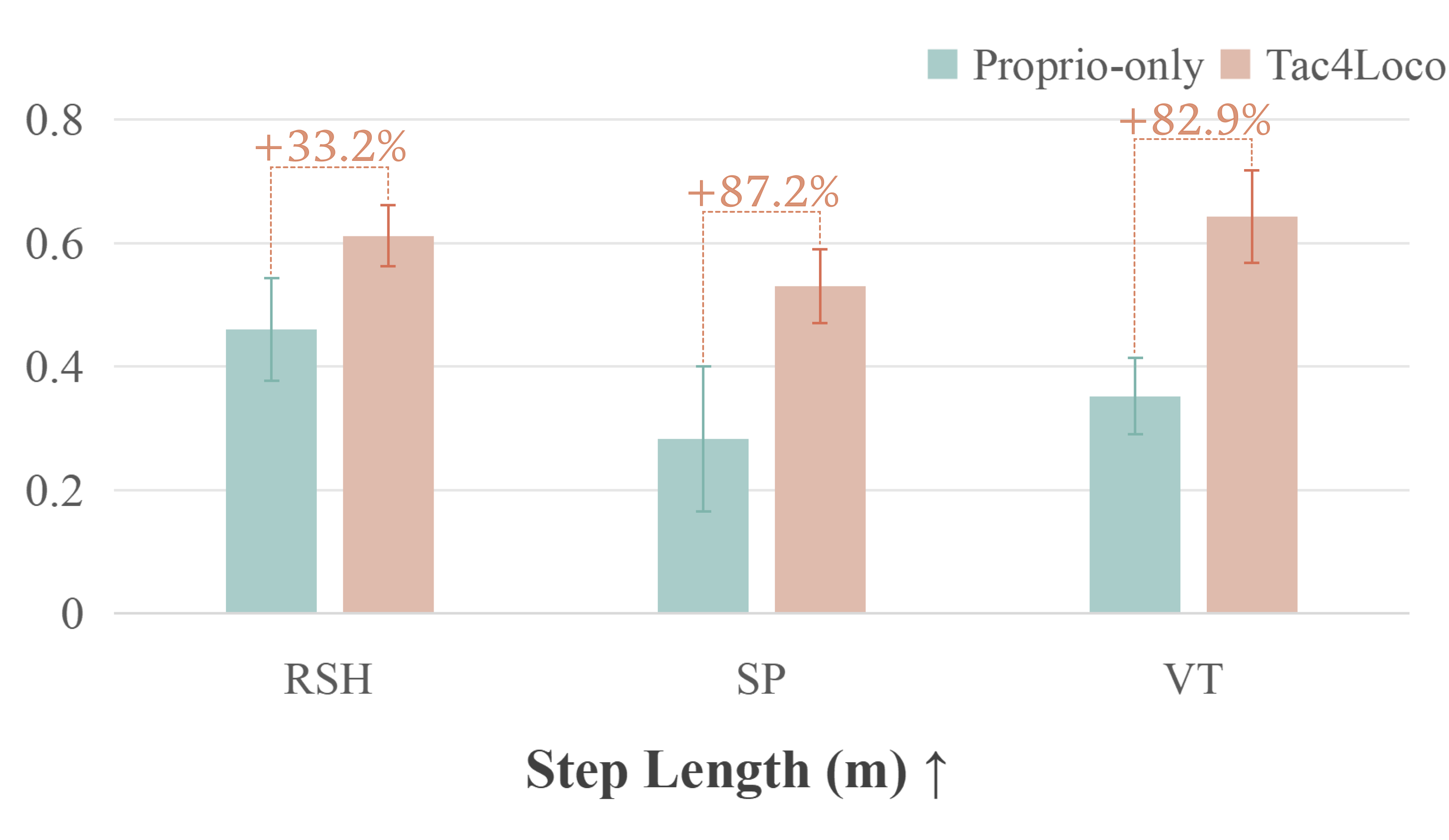}
    \caption{Step length under challenging support conditions.
Error bars denote standard deviation.}
    \label{fig:step_length}
\end{figure}

As shown in Fig.~\ref{fig:step_length}, Tac4Loco exhibits longer step lengths on random support-height terrain, slopes, and the V-trench.
When interpreted jointly with survival and velocity-tracking performance in Table~I, these results indicate less disrupted progression under challenging support conditions.

\subsection{Ablation Study}

Table~\ref{tab:ablation} isolates the contributions of the dual-branch plantar pressure encoder and the contact-conditioned terrain-orientation feature. The reported terrain--metric combinations are selected to expose the clearest behavioral effect of each component rather than repeat the complete benchmark.

Removing the terrain-orientation feature primarily degrades locomotion under continuously varying or inclined support. The increase in velocity-tracking error on gently undulating terrain, together with the reduction in slope survival and effective step progression, indicates that this feature provides a compact physical cue for adapting the gait to the orientation of the realized support surface. The pressure maps still describe the contact distribution, but without the summarized orientation cue the policy has greater difficulty converting that information into sustained slope-adaptive motion.

Removing the pressure encoder produces its clearest degradation under localized, asymmetric, and discontinuously changing support. In the V-trench, the ablated policy shows lower survival and greater lateral drift, consistent with failures caused by gradually deviating from the intended traversal direction. On random support-height terrain, the smaller reduction in survival and the increase in touchdown-interval variation indicate weaker regulation of changing support patterns. These results show that the topology-preserving spatial representation and the temporal pressure history supply contact information that cannot be recovered from the compact terrain-orientation feature alone.

\begin{table}[t]
\centering
\caption{Open-terrain drift angle under translational commands.}
\label{tab:drift_angle}

\renewcommand{\arraystretch}{1.10}
\small

\begin{tabular}{@{}lcc@{}}
\toprule
Terrain
& Proprio-only
& Tac4Loco \\
\midrule

FL
& 1.434 $\pm$ 0.065
& \textbf{0.495 $\pm$ 0.048} \\

GU
& 1.333 $\pm$ 0.520
& \textbf{0.551 $\pm$ 0.258} \\

RSH
& 1.095 $\pm$ 0.482
& \textbf{0.545 $\pm$ 0.337} \\

\bottomrule
\end{tabular}

\begin{flushleft}
\footnotesize
\textit{Note.}
Drift angle is reported in radians and averaged over the four translational commands.
Lower is better.
\end{flushleft}

\end{table}

\begin{table*}
\centering
\caption{Ablation study of Tac4Loco components.}
\label{tab:ablation}
\resizebox{\textwidth}{!}{
\begin{tabular}{lccccccc}
\toprule
\multirow{2}{*}{Method}
& \multicolumn{4}{c}{Task Performance}
& \multicolumn{3}{c}{Gait and Motion Metrics} \\
\cmidrule(lr){2-5}
\cmidrule(lr){6-8}
& \shortstack{GU\\Lin. Vel. Err. (m/s) $\downarrow$}
& \shortstack{RSH\\Surv. (\%) $\uparrow$}
& \shortstack{SP\\Surv. (\%) $\uparrow$}
& \shortstack{VT\\Surv. (\%) $\uparrow$}
& \shortstack{RSH\\Step Interval CV (-) $\downarrow$}
& \shortstack{SP\\Step Len. (m)}
& \shortstack{VT\\Lateral Drift (m) $\downarrow$} \\
\midrule
w/o Pressure Enc.
& 0.158 $\pm$ 0.041
& 90.6
& 73.4
& 68.8
& 0.228 $\pm$ 0.109
& 0.421 $\pm$ 0.055
& 0.194 $\pm$ 0.077 \\
w/o Terrain Est.
& 0.228 $\pm$ 0.061
& 94.9
& 42.2
& 96.9
& 0.201 $\pm$ 0.102
& 0.267 $\pm$ 0.063
& 0.067 $\pm$ 0.029 \\
Tac4Loco
& \textbf{0.140 $\pm$ 0.030}
& \textbf{96.5}
& \textbf{77.9}
& \textbf{100.0}
& \textbf{0.195 $\pm$ 0.098}
& \textbf{0.452 $\pm$ 0.047}
& \textbf{0.059 $\pm$ 0.023} \\
\bottomrule
\end{tabular}}
\begin{flushleft}
\footnotesize
\textit{Note.} Terrain abbreviations: GU = gently undulating, RSH = random support-height, SP = slopes, VT = V-trench.
Values with $\pm$ denote mean $\pm$ standard deviation over pooled episode-level values.
Selected terrain--metric pairs highlight the clearest component effects; step length is interpreted jointly with survival rate.
\end{flushleft}
\end{table*}

The two components therefore play complementary roles. The local terrain-orientation feature mainly supports velocity regulation and effective progression over continuously varying support directions, whereas the dual-branch pressure encoder preserves the spatial and temporal contact structure required for heading regulation and support-transition control under asymmetric or locally uncertain contact. Their combination produces the terrain-adaptive capabilities examined again in the real-world ramp, trench, terrain-edge, and rigid-support transition experiments.

\begin{figure*}[t]
    \centering
    \vspace{6pt}
    \includegraphics[width=1\linewidth]{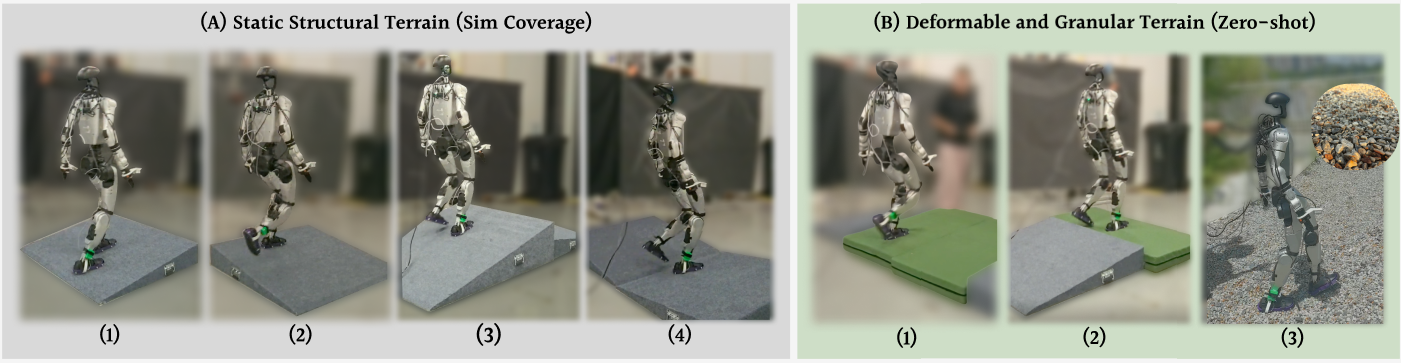}
    \caption{Real-world deployment validation of Tac4Loco.
    \textbf{(A)~Static structural terrain} (covered by simulation):
    (1)~forward traversal over a $9^\circ$ ramp and transition from its upper platform to lower flat ground;
    (2)~lateral ascent onto a $9^\circ$ ramp;
    (3)~transition from a $15^\circ$ ascending ramp to a $9^\circ$ descending ramp;
    (4)~traversal of a V-trench formed by two $9^\circ$ ramps.
    \textbf{(B)~Deformable and granular terrain} (zero-shot):
    (1)~transition from rigid flat ground onto a compliant foam platform;
    (2)~transition from a $9^\circ$ ramp onto the compliant platform;
    (3)~gravel road.
}
    \label{fig:real_world}
\end{figure*}

\subsection{Real-World Deployment Validation}
Exact scene-level correspondence between simulation and physical deployment is impractical because terrain geometry, contact location, and realized pressure distributions cannot be reproduced identically. We therefore organize the real-world validation around recurring foot--ground interaction characteristics rather than one-to-one terrain matching.

We deployed Tac4Loco on a Unitree G1 humanoid robot equipped
with bilateral plantar pressure insoles. The physical measurements
were mapped into the same topology-preserving ordinal pressure
representation used in simulation, preserving the physical meaning
and spatial organization of plantar support across training and
deployment. Both the locomotion controller and the pressure measurements operated at 50 Hz. 
To account for the remaining sensing gap, pressure-specific augmentation was applied
during training to model measurement variation, temporal distortion,
and temporary sensing failures. These perturbations were disabled
during deployment, where the calibrated physical measurements were
directly provided to the policy.

\subsubsection{Real-World Validation Results}
The physical experiments were designed to validate Tac4Loco under contact conditions that are difficult to reproduce fully in simulation.
Table~\ref{tab:real_deployment} summarizes the tested configurations and their primary support challenges, together with real-world completion results for Tac4Loco and the proprioception-only baseline where evaluated.
Completion is stricter than the survival metric used in simulation, as it requires the robot to traverse the prescribed terrain without falling.
Tac4Loco achieved consistently higher completion than the proprioception-only baseline across the evaluated comparisons, particularly under inclined, asymmetric, transitional, and compliant support conditions.
Representative motion sequences and the corresponding bilateral plantar pressure maps at key contact states are shown in Fig.~\ref{fig:real_world}.

\begin{table}[h]
\centering
\caption{Real-world deployment comparison of Tac4Loco and the proprioception-only baseline.}
\label{tab:real_deployment}
\scriptsize
\begin{tabular*}{\columnwidth}{@{\extracolsep{\fill}}llcc@{}}
\toprule
\multicolumn{1}{c}{\multirow{2}{*}{Configuration}}
& \multicolumn{1}{c}{\multirow{2}{*}{Support challenge}}
& \multicolumn{2}{c}{Completion} \\
\cmidrule(lr){3-4}
&
& Proprio-only
& Tac4Loco \\
\midrule
$9^\circ$ ramp edge
& Partial support / drop
& 7/10
& 10/10 \\
Lateral $9^\circ$ ascent
& Asymmetric support
& 1/10
& 8/10 \\
$15^\circ$ up + $9^\circ$ down
& Slope transition
& 0/10
& 10/10 \\
$9^\circ$ V-trench
& Edge support
& --
& 10/10 \\
Flat $\to$ foam
& Compliant transition
& 0/10
& 7/10 \\
Ramp $\to$ foam
& Slope + compliance
& 4/10
& 10/10 \\
\bottomrule
\end{tabular*}
\begin{flushleft}
\footnotesize
\textit{Note.} The proprioception-only baseline was not further deployed on the V-trench because it did not exhibit sustained forward progression on this terrain during curriculum training.
\end{flushleft}
\end{table}

\subsubsection{Qualitative Observations}
\textbf{Group A: static structural terrains.}
We test our proposed controller and the proprioception-only baseline across terrain configurations with abrupt support-height changes, including the upper-platform to flat transition of the $9^\circ$ ramp and the junction between the $15^\circ$ ascent and $9^\circ$ descent.
Without terrain geometry preview, these controllers cannot identify terrain boundaries in advance, often causing the foot to land with only partial support.

The proprioception-only baseline exhibits distinct failure modes as the support condition becomes more demanding.
On the $9^\circ$ ramp-edge transition, a low forward-velocity command often prevents the baseline from climbing onto the ramp, whereas a higher command enables ascent but produces an abrupt drop at the platform-to-flat height discontinuity. 
On the steeper $15^\circ$ ascent, the baseline cannot climb the ramp at all. 
Moreover, in lateral ascent tasks, the baseline generally remains upright, but struggles in tracking the commanded velocity direction and gradually drifts downslope back onto the surrounding flat ground, rather than walk across the ramp laterally.

In contrast, Tac4Loco achieves smooth downward transitions and higher completion rates, as shown in the first four rows of Table~\ref{tab:real_deployment}. This improvement stems from localized plantar pressure sensing, which provides vital haptic feedback immediately following touchdown and enables rapid load redistribution and reactive balance adjustment after support loss.

\textbf{Group B: zero-shot deformable and granular terrains.}
To validate the zero-shot transferability of Tac4Loco, we also test it on deformable (foam) and granular (gravel) terrains absent from simulation training.
On the foam, the surface deforms after touchdown, causing continuous variation of the support force and contact area. This characteristic is particularly challenging for humanoid locomotion, and is difficult to reproduce in large-scale parallel reinforcement learning simulation platforms.

Under the compliant setting, the proprioception-only baseline cannot reliably infer plantar support from proprioception (especially leg joint states) as it can on solid ground. When the foot partially lands on the foam, the concentrated local contact force induces large surface deformation, raising the probability of falling and yielding substantially lower completion rates.
In contrast, Tac4Loco achieves higher completion rates as shown in the last two rows of Table~\ref{tab:real_deployment}. This improvement stems from explicit multi-array plantar pressure modeling, which provides direct post-contact feedback of the evolving support topology, allowing the policy to judge the support state from contact forces, stabilize the center of mass, and thereby adapt across terrains of different properties.
On gravel, shifting local support points produce fragmented and time-varying plantar loading. We conducted approximately five minutes of continuous testing on the gravel road, during which Tac4Loco maintained robustness without additional training. 

Overall, these experiments demonstrate zero-shot transfer to support conditions whose contact geometry continues to evolve after touchdown through either compliant deformation or granular rearrangement.

%% file: Sections/6_Conclusion.tex
\section{Conclusion}

In this paper, we present Tac4Loco, a tactile-perceptive humanoid locomotion framework that incorporates multi-array plantar pressure as direct contact-posterior feedback. 
By leveraging a dual-branch encoder to extract topology-preserving spatiotemporal pressure representations alongside contact-conditioned terrain-orientation cues,
Tac4Loco enables contact-aware policy learning and sim-to-real deployment. Simulation experiments demonstrate improved velocity tracking, locomotion robustness, and support adaptation over a proprioception-only baseline, while ablations reveal the complementary benefits of pressure perception and terrain-orientation cues under diverse contact conditions. Real-world experiments further demonstrate deployment across ramps, trenches, terrain edges, changing rigid supports, and unseen compliant surfaces. 
Future work will explore the multi-modal integration of plantar pressure feedback with exteroceptive visual sensing to unify terrain anticipation with post-contact reactivity.

%% file: Sections/9_Bibliography.tex
\balance

\bibliographystyle{ieeetr}
\bibliography{root}